\documentclass[runningheads]{llncs}

\usepackage{iciap}

\usepackage{iciapabbrv}
\newcommand{\red}[1]{\textcolor{black}{#1}}

\usepackage{graphicx}
\usepackage{booktabs}
\usepackage{graphicx}
\usepackage{amsmath}
\usepackage{amssymb}
\usepackage{tabularx}
\usepackage{booktabs} 
\usepackage{siunitx}
\usepackage{xcolor}
\usepackage{enumitem}
\usepackage{comment}

\usepackage[accsupp]{axessibility}  %

\usepackage{hyperref}

\usepackage{orcidlink}

\title{Segment Anything for Satellite Imagery: A Strong Baseline and a Regional Dataset for Automatic Field Delineation}

\newcommand*{\affmark}[1][*]{\textsuperscript{#1}}

\begin{document}

\title{Segment Anything for Satellite Imagery: A Strong Baseline and a Regional Dataset for Automatic Field Delineation} 

\titlerunning{Segment Anything for Satellite Imagery}

\author{Carmelo Scribano\affmark[1, 3]\orcidlink{0000-0003-1006-7826}, Elena Govi\affmark[1]\orcidlink{0000-0002-4167-3741}, Paolo Bertellini\affmark[2], Simone Parisi\affmark[2],\\ Giorgia Franchini\affmark[1]\orcidlink{0000-0001-9082-8087} and Marko Bertogna\affmark[1]\orcidlink{0000-0003-2115-4853}
}
\authorrunning{C.~Scribano, E.~Govi, et al.}

\institute{University of Modena and Reggio Emilia, Italy. \\
\email{\{name.surname\}@unimore.it} \and
ABACO Group SpA, Mantova, Italy.\\
\email{\{p.bertellini, s.parisi\}@abacogroup.eu}\and
Institute of Informatics and Telematics, National Research Council, Italy.\\
}

\maketitle

\begin{abstract}
Accurate mapping of agricultural field boundaries is essential for the efficient operation of agriculture. 
Automatic extraction from high-resolution satellite imagery, supported by computer vision techniques, can avoid costly ground surveys. In this paper, we present a pipeline for field delineation based on the Segment Anything Model (SAM), introducing a fine-tuning strategy to adapt SAM to this task. In addition to using published datasets, we describe a method for acquiring a complementary regional dataset that covers areas beyond current sources. Extensive experiments assess segmentation accuracy and evaluate the generalization capabilities. Our approach provides a robust baseline for automated field delineation. The new regional dataset, known as ERAS, is now publicly available. \footnote{\url{https://github.com/cscribano/ERAS-dataset}}.

\end{abstract}

\section{Introduction}

\label{sec:intro}

\begin{figure}[ht]
    \centering
    \begin{subfigure}[c]{0.21\textwidth} %
        \includegraphics[width=\textwidth, clip]{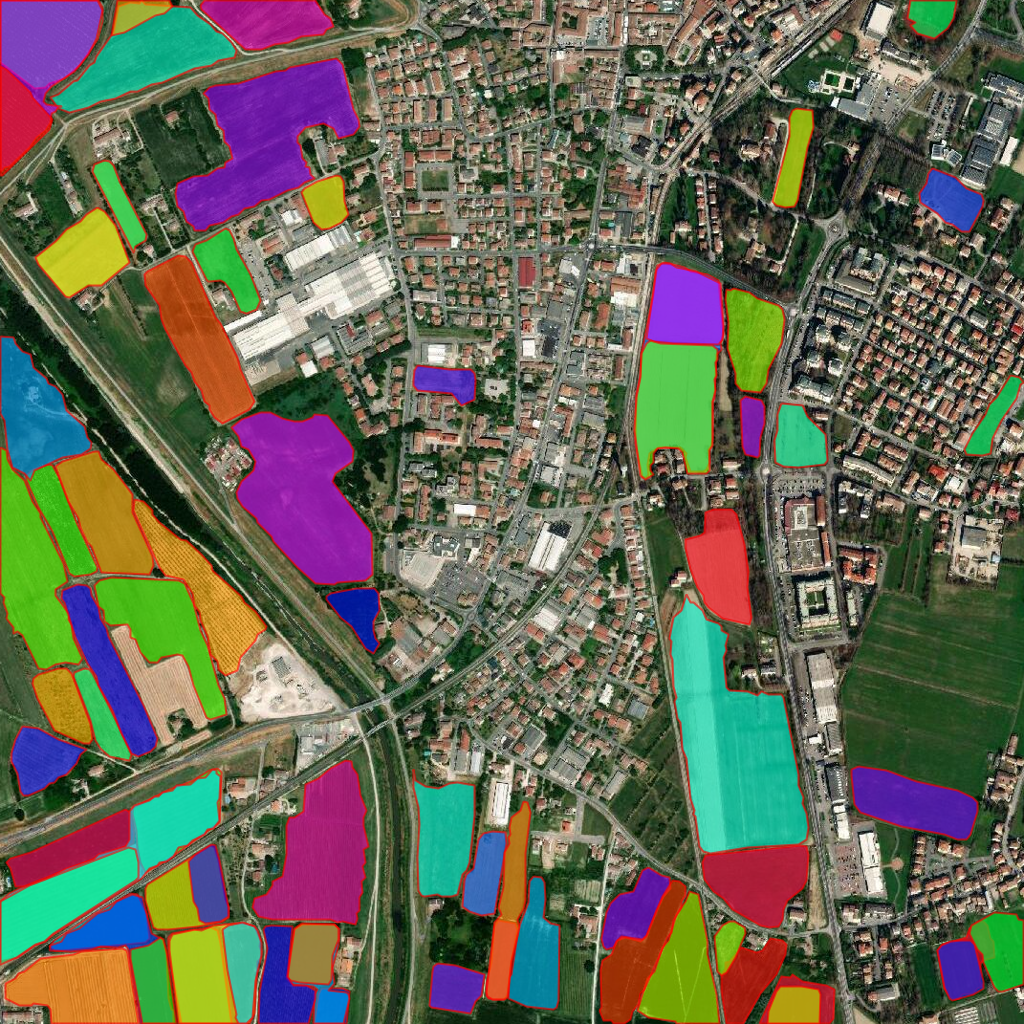}
    \end{subfigure}
    \begin{subfigure}[c]{0.21\textwidth}%
        \includegraphics[width=\textwidth, clip]{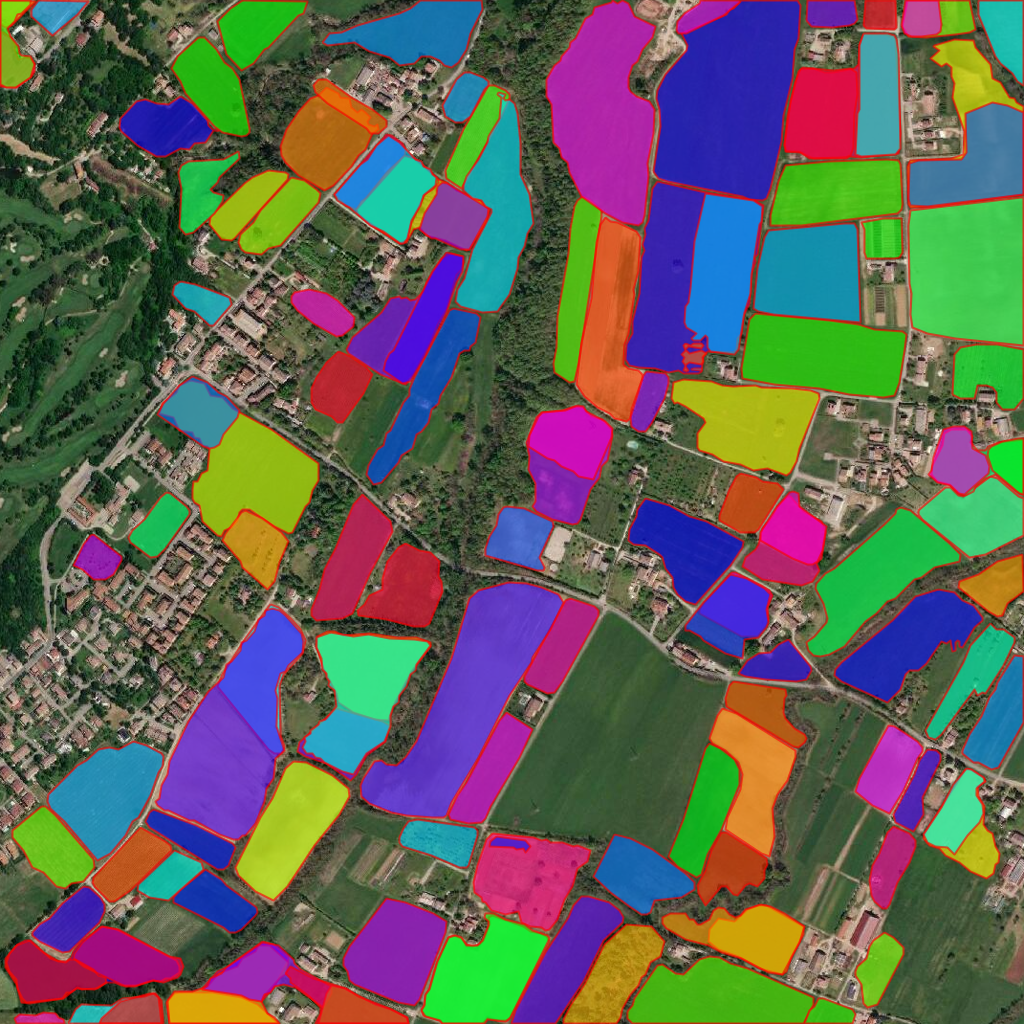}
    \end{subfigure}
    \begin{subfigure}[c]{0.21\textwidth} %
        \includegraphics[width=\textwidth, clip]{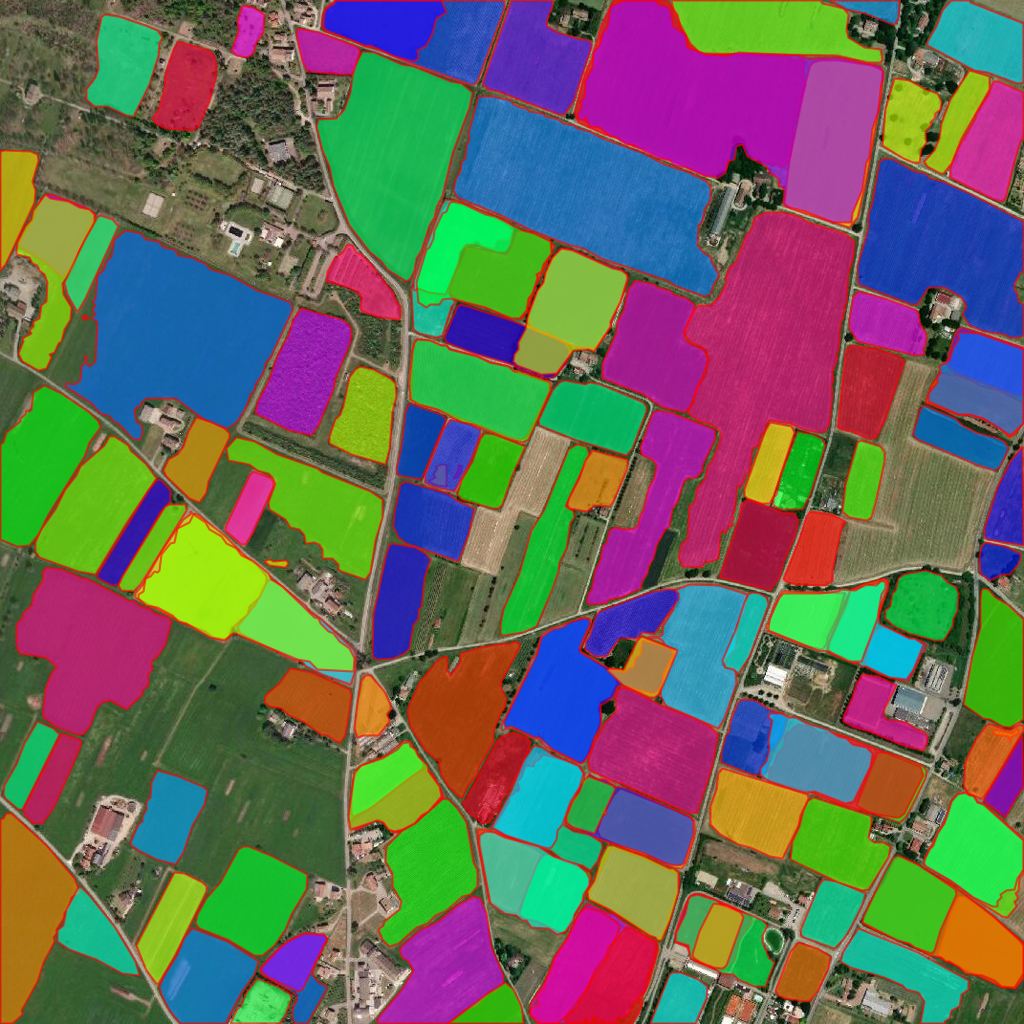}
    \end{subfigure}
    \begin{subfigure}[c]{0.21\textwidth} %
        \includegraphics[width=\textwidth, clip]{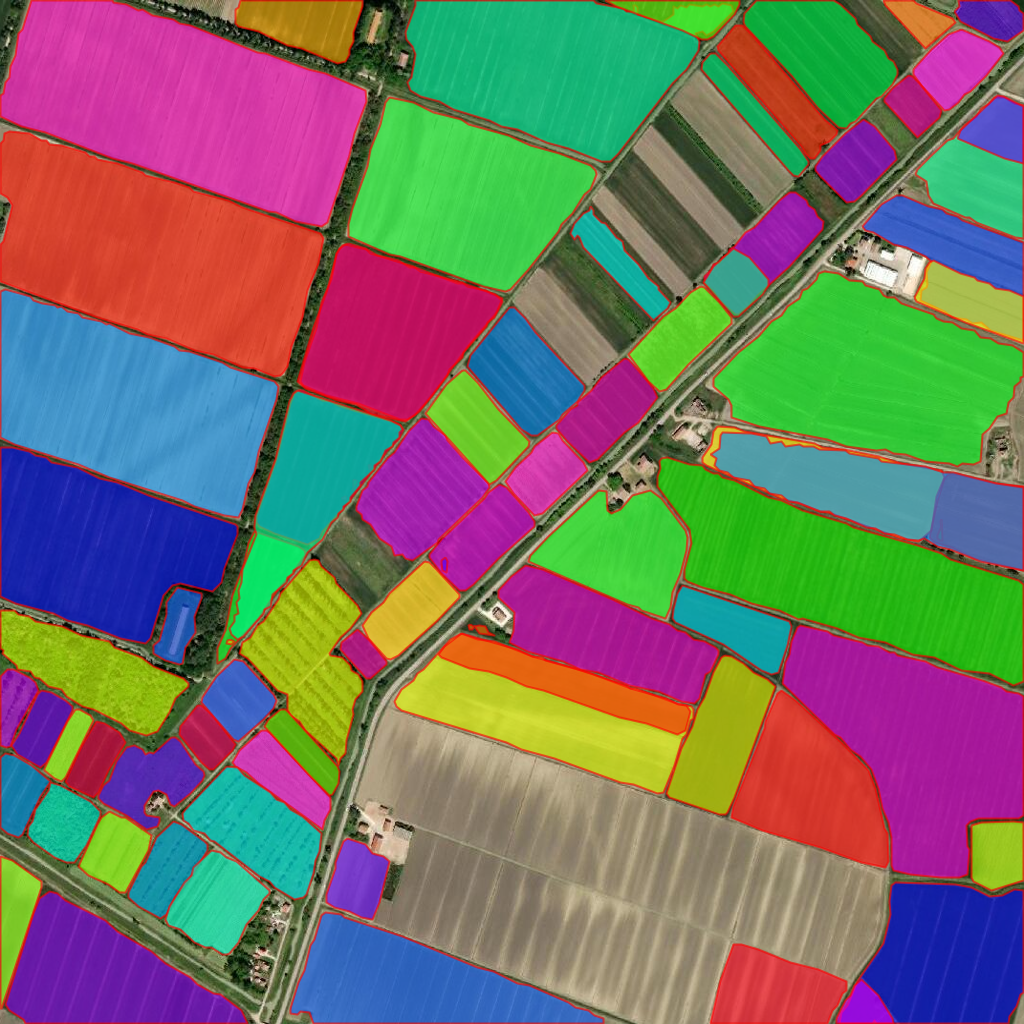}
    \end{subfigure}
    \caption{Example predictions from finetuned SAM on the proposed dataset. Additional visualizations are provided in the supplementary material.}
    \label{fig:gt}
\end{figure}

\noindent

Accurately mapping the geographical boundaries of cultivated fields is essential for agricultural operations in several ways: enhancing land planning and management, improving resource allocation, supporting large-scale crop monitoring, and ensuring accurate subsidy distribution based on cultivated areas. Traditional ground-based GPS surveys are slow and resource-intensive, while manual delineation from high-resolution satellite imagery remains labor-intensive and prone to errors at scale. Automating this mapping process using computer vision techniques offers an attractive solution to reduce effort, costs, and inaccuracies. The goal is to automatically extract the boundaries of all visible fields from a satellite image, where a field is a portion of land designated for crop cultivation. Among other options, this problem can be cast as a subset of the instance-level segmentation task of agricultural parcels from satellite imagery. This choice allows leveraging a vast range of existing techniques from the realm of image segmentation, among which models based on Deep Learning stand out. Training a powerful instance segmentation model from scratch for this task presents several challenges. The obtainable datasets are typically too small, and unannotated objects in the ground truth can hinder the training process. To address these challenges, we focus on fine-tuning the Segment Anything Model (SAM), which has the potential to mitigate these limitations. SAM \cite{kirillov2023segment} represents a relevant paradigm shift in the segmentation domain. SAM is pretrained on the large-scale dataset SA-1B, which consists of 11M images and 1.1B mask annotations, achieving strong zero-shot generalization abilities in several domains. These characteristics have made SAM an extremely popular choice for a wide set of segmentation problems, but its use for field delineation has not yet been explored in depth. Our early experiments reveal subpar zero-shot performance in our domain. In this paper, we propose a fine-tuning strategy for SAM based on the Low-Rank adaptation (LoRA) methodology \cite{hu2021lora}. By leveraging this approach, we adapt SAM to our specific use case, achieving strong segmentation performance even with a limited number of training examples.\\

To enable automated mapping approaches, access to high-quality satellite imagery is crucial. A widely used source is the European Space Agency’s (ESA) Sentinel-2 missions, which provide Red, Green, Blue (RGB) and Near-Infrared (NIR) bands at a 10-meter spatial resolution. For field delineation tasks, the AI4Boundaries (AI4B) dataset \cite{d2023ai4boundaries}, based on RGB Sentinel-2 imagery, offers an excellent foundation for experimental analysis. In addition, we collect ERAgriSeg (ERAS), a new, original dataset specific to the Emilia-Romagna (ER) region of northern Italy. This dataset serves to evaluate the generalization capabilities of models fine-tuned on AI4B and to assess fine-tuning performance when working with fewer examples and a smaller spatial extent. Alongside Sentinel-2 imagery, our dataset includes a high-resolution variant (2.5 meters per pixel) acquired from a commercial satellite constellation. Both versions cover the same geographical areas and share identical segmentation masks; however, due to licensing restrictions, we are only able to release the Sentinel-2 images. The experiments presented in this paper make use of both image modes, offering insights into the influence of spatial resolution on the field delineation task.

\noindent The main contributions of this work can be summarized as:
\begin{itemize}
    \item A study on the applicability of SAM to the field delineation problem, heavily investigating the option of fine-tuning with LoRA and evaluating its implications;
    \item The acquisition of a dataset for a geographical area (Emilia Romagna, Italy) not covered by existing datasets, pushing the limits of the proposed method to new geographical areas, relying on a limited number of training samples.
    \item An analysis of the temporal generalization capabilities of the proposed method.\\
\end{itemize}
The paper is structured as follows: \Cref{sec:related} reviews key computer vision approaches for satellite imagery and the field delineation problem. \Cref{sec:data} describes the usable datasets, including the proposed one. \Cref{sec:method} outlines our adaptation of SAM for field delineation. \Cref{sec:results} presents experimental results, and \Cref{sec:concl} concludes with future directions. %

\section{Related Work}
\label{sec:related}
\subsection{Computer Vision for Earth Observation}

Machine and deep learning have notably advanced Earth Observation (EO), particularly in satellite image classification, object detection, and segmentation. Image classification assigns labels based on spectral, spatial, and temporal features, supporting various geospatial applications. Notable examples include EuroSAT \cite{helber2019eurosat}, a Sentinel-2-based benchmark for land use and land cover classification, and crop classification methods \cite{garnot2019time, bertellini2023binary}. Automatic field delineation, formulated as an instance segmentation task, involves identifying field pixels and distinguishing individual fields. This task is challenged by high-frequency noise, missing pixel values, seasonal dynamics, weather variability, and atmospheric distortions affecting image quality. Before the rise of deep learning, several traditional methods were proposed for similar problems \cite{mueller2004edge, evans2002segmenting}. The introduction of Convolutional Neural Networks (CNNs) and Transformer-based architectures has since led to substantial performance improvements. ResUNet-based models, for example, have been explored in \cite{waldner2020deep, diakogiannis2020resunet, waldner2021detect}, though the absence of publicly available code and data limits reproducibility and comparison. Other notable contributions include \cite{garcia2017machine, garcia2019deep, Aung_2020_CVPR_Workshops, waldner2020deep, sun2024enhancing, d2023ai4boundaries}. Field segmentation remains an active research area, with recent advances yet considerable room for further improvement.

\subsection{Segment Anything}
Segment Anything \cite{kirillov2023segment} (SAM) is a modern foundation model for instance-level segmentation. SAM predicts instance-level segmentation masks for a given image based on input prompts, such as points, bounding boxes, or a coarse segmentation mask. 
Recent works demonstrate remarkable results in different domains, ranging from medical image segmentation \cite{wu2023medical}, aerial land cover classification \cite{xue2024adapting}, and satellite images. SAM's potential for satellite imagery analysis has been explored in prior work. The work in \cite{osco2023segment} evaluated SAM across multi-scale remote sensing datasets, developing an automated method that combines text prompts with one-shot training. Similarly, \cite{wang2024samrs} introduced SAMRS, a large-scale remote sensing dataset generated using SAM for segmentation and classification labeling. In our study, SAM is used to extract instance-level field masks, from which field boundaries can be easily derived. LoRA, a popular parameter-efficient fine-tuning approach, is proposed to adapt SAM for satellite imagery.

\section{Field Delineation Datasets}
\label{sec:data}

\subsection{Existing Datasets}\label{subsec:existing}

Most segmentation datasets for earth-observation, are designed for the segmentation of roads, buildings, and a variety of geographical features of interest. Some specific datasets have been published so far for the segmentation of cultivated fields. Among those, AI4SmallFarms \cite{persello2023ai4smallfarms} is a small dataset composed of $62$ samples only. %
To the best of our knowledge, the main datasets suitable for the task of field delineation with a segmentation approach are Eurocrops \cite{schneider2023eurocrops} and AI4Boundaries (AI4B) \cite{d2023ai4boundaries}. The first, Eurocrops, covers 13 EU countries, with labels obtained from farmers' self-declarations. The dataset covers the years 2015-2022, but for each year, only a subset of the 13 countries is covered. %
Except for a small subset, image-mask pairs are not yet provided ready for segmentation, representing this dataset's main limitation. Only the geospatial vector data for all plots is provided, so it would be necessary to define the tiles for segmentation and obtain the associated satellite image.

\noindent The second dataset, AI4B, covers seven regions in the EU zone. Unlike the first dataset, this one provides preprocessed tiles that include both satellite images and their corresponding plot boundaries, derived from self-declarations. The satellite images, sourced from ESA’s Sentinel-2 missions, have a spatial resolution of 10 meters per pixel and an extent of $256\times256$ pixels. Additionally, the dataset includes aerial orthophotos with a 1-meter resolution. The authors propose an image aggregation algorithm that combines monthly data to produce cloud-free composites. In total, the dataset contains 7,831 samples, with over 2.5 million unique parcels. Monthly aggregate images are provided for March through August 2019.

\subsection{Proposed Dataset}\label{subsec:er4seg}

\begin{table}[h]
    \caption{Comparison of datasets for field delineation}
    \centering
    \scalebox{0.8}{
    \begin{tabular}{ccccc}
    \toprule
    \textbf{Dataset} & \textbf{n-Samples}  & \textbf{Resolution} &\textbf{Countries} & \textbf{Years}\\
    \midrule
    AI4SmallFarms & $62$ & $256\times 256 $ & VN, KH & 2021 \\ %
    
      Eurocrops &  - & - & 17 EU & 2015 - 2022\\
      
       AI4B   & \num{7831} & $256\times256$& \begin{tabular}{c}
            AT, ES, FR  \\
           SE, NL, SI, LU
       \end{tabular} & 2019\\
       
      \textbf{ERAS (S2)} & \num{14968} & $256\times256$ & IT & 2023-2024(Q1)\\
      \textbf{ERAS (HR)} & \num{3742} & $1024\times1024$ & IT & $\sim$2023\\ 
      \bottomrule
    \end{tabular}
    }
    \label{tab:dataset}
    
\end{table}

This work derives from a study of the application of field delineation techniques to the Italian territory, and in particular the Emilia-Romagna (ER) region (northern Italy), characterized by a vast flatland and with a high agricultural surface area by Italian standards (third overall by surface area). It is administered in 9 provinces (Bologna (BO), Modena (MO), Reggio-Emilia (RE), Ferrara (FE), Rimini (RN), Piacenza (PC), Ravenna (RA), Parma (PR), Forlì-Cesena (FC)). The reasons for aspiring to a region-specific dataset are many, first, because this region, and Italy in general, is not covered by existing datasets (AI4B, Eurocrops, and AI4SmallFarms). 
It is important to be able to evaluate the generalization ability of fine-tuned SAM on publicly available datasets to regions not present in the original training set. Second, we want to test the possibility of further improving the results by employing only a small set of labeled images obtainable from a regional ad-hoc dataset. Finally, we had the opportunity to acquire higher-resolution images for the region of interest, which allowed the investigation of the impact of spatial resolution. \red{Experiments conducted on both datasets at varying resolutions allowed for a novel comparison that had not been explored before.} We dub the proposed dataset as ERAgriSeg, or just \textbf{ERAS} for short.

\subsubsection{Acquisition and Preprocessing}

To maintain consistency with AI4B, the segmentation images are defined as square tiles, each covering a 2560-meter by 2560-meter area. Tiles are accompanied by segmentation masks for the parcels and the corresponding satellite observations. The Geospatial data for parcel boundaries, used to define the ground-truth segmentation masks, were obtained from self-reported statements submitted by farmers to the administration in 2023. The tile extraction algorithm is applied to each province. First, a set of candidate tile centroids is defined using a k-means clustering algorithm applied to the area bounded by the ground-truth edges. The number $K$ of clusters is equal to 110\% of the total area of the fields, divided by the extent tile extent (6.5536$km^2$). A tile is defined around each centroid. To generate the mask, the parcel boundaries are preprocessed by excluding certain categories (e.g., non-agricultural land, landscape elements) and malformed parcels.

\noindent As mentioned, the acquired dataset consists of 2 modalities, the first corresponds with observations from ESA's Sentinel-2 satellites, for the second, high-resolution mode, instead, the images are obtained with a license from Maxar Technologies. We use RGB Sentinel-2 images at 10m resolution. Under optimal conditions, the combined revisit time of Sentinel-2A and 2B satellites is approximately 5 days at the equator. Instead of using raw observations, which are affected by cloud cover and other defects, we utilize the cloudless Sentinel-2 Level 3 Quarterly Mosaics (L3 mosaics) provided by ESA. These mosaics are generated by aggregating all observations over 3 months, starting from January 2022. For each pixel, invalid observations are discarded, and the pixel value is set to the average of the first quartile of the remaining valid observations for each band. For the year 2023, all four quarterly aggregates are available. Ground-truth boundaries for 2024 and corresponding Sentinel-2 images for the first quarter are also available, but this subset is excluded from the main training set. In \Cref{subsec:y2024}, we use it to evaluate temporal generalization to 2024 without further fine-tuning. The second data source provides high-resolution images at 2.5m per pixel. These images are pre-processed to remove clouds and defects, but due to their commercial nature, specific details are not disclosed. Unlike the Sentinel-2 data, acquisitions are not periodic, so for each tile, we gather the most recent available observation, primarily covering the year 2023 (denoted as $\sim$2023 in \Cref{tab:dataset}). For both modalities, unless otherwise specified, we isolate all observations for the province of Reggio Emilia (RE) as the validation set. Hereafter, we refer to the version using Sentinel-2 images as the `S2' version and the high-resolution Maxar images as the `HR' version.

\begin{figure*}[h!]
    \centering
    \includegraphics[width=0.8\textwidth, clip]{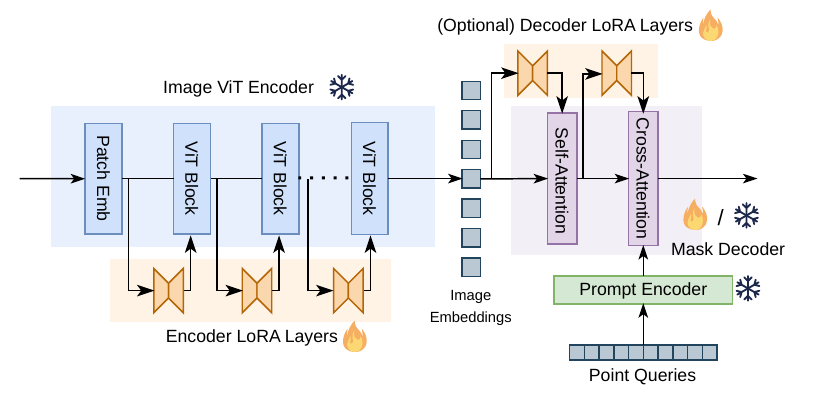}
    \caption{Schematic of SAM architecture in fine-tuning phase. The snowflake indicates that the layer is being frozen, and the flame indicates that it is being trained. For the mask decoder, we experiment with three options (a):  frozen, (b): full finetuning, and (c): frozen and adding LoRA layers.}
    \label{fig:samLoRA}
\end{figure*}
\vspace{-20pt}

\section{Proposed Methodology}
\label{sec:method}

\paragraph{Data Preparation}
The tiles obtained from Sentinel-2 have a resolution of $(256\times256)$ pixels, unlike the CNNs; however, SAM's transformer encoder \red{does not tolerate changes in the input size unless re-trained}. As a result, it is necessary to rescale the images to the native size of $(1024\times1024)$, the high-resolution images in the ERAS dataset, on the other hand, already come at the correct resolution. %
For both datasets, we use data augmentation during the training phase.
\subsection{SAM Finetuning}
\label{subsec:sft}

\paragraph{Low-Rank Adaptation}

Parameter-Efficient Fine-Tuning (PEFT) methods enable training with minimal parameter updates. LoRA \cite{hu2021lora}, a popular PEFT technique, stems from the intuition that a matrix of weights $W_0$ after fine-tuning will result in $W_T=W_0+\Delta W$, assuming that $\Delta W$ is low rank it is possible to decompose it as the product of two smaller matrices $A$ and $B$. Accordingly, the calculation of the result $y_0$ obtained by applying $W_T$ to an input $x_0$ can be expressed as (omitting the bias term for simplicity):
   $ y_0 = W_0x + \Delta Wx = W_0x + ABx,  $
 with $W_0\in \mathcal{R}^{d\times k} $, $A \in \mathcal{R}^{r\times k }$ and $B\in \mathcal{R}^{d\times r}$, where $r$ is typically referred to as LoRA Rank. Given that the rank $r<< \min(d,k)$ is lower than $d$ and $k$, the result is a substantial reduction in the parameters learned by the model. The LoRA layer is applied to the projection layers in the attention heads; after training, $\Delta W$ can be computed and added to the original weights $W$. %

\noindent The ViT backbone of SAM is compatible with LoRA and similar methods. Some existing works \cite{xue2024adapting, wu2023medical} apply similar PEFT techniques to successfully specialize SAM on particular domains. Driven by these observations, we propose a LoRA-based fine-tuning scheme for the extension of SAM to field segmentation. %

\paragraph{Fine-tuning Strategy}
The complete SAM model, schematized in \Cref{fig:samLoRA}, consists of three elements: the image encoder maps the input image into a set of image tokens. The prompt encoder is a simple embedding layer that maps the prompt, in our case a set of $N$ points, into a set of $N$ spatial embeddings. Finally, the mask decoder is a transformer decoder that applies cross-attention to produce a set of $P$ instance-level masks from the image tokens and the prompt embeddings. In our pipeline, LoRA is applied to the image encoder, while the prompt encoder is completely frozen for all experiments. Regarding the mask decoder, we consider three different options: %
  (i) freeze all weights,
  (ii) train all weights at full rank, without introducing LoRA layers,
  (iii) apply LoRA to self-attention and cross-attention layers, freezing all other parameters.
In the next chapter, we discuss and compare the results obtained in these three possible configurations. 

\subsection{Prompt Definition}
In the simplest case, the prompt is a single point $P=(x_p, y_p)$, for which SAM will produce an instance prediction. 
\red{Otherwise}, SAM accepts as a prompt a set of points for each instance, for each point a binary label is also provided, indicating whether it is positive (the point belongs to the ground-truth mask) or negative (the point is outside the mask). The choice of prompt type during training will affect performance. In this paper, we compare two choices: %
(i) provide, for each instance, a single randomly sampled positive point from the ground-truth mask, (ii) simultaneously provide 4 positive and 2 negative points, sampled near the edges of the mask.
\red{This choice affects only the training phase, while during evaluation and test, the prompts are obtained by a static grid, as detailed in \Cref{subsec:setup}}.

\section{Experiments and Evaluation}
\label{sec:results}
To evaluate the effectiveness of SAM on the field delineation task, we begin by assessing zero-shot performance on AI4B and the proposed dataset in \cref{subsec:zeroshot}. Subsequently, in \cref{subsec:sam_ft}, we focus on fine-tuning SAM, comparing the results with a more conservative model in \cref{subsec:rcnn_baseline}. In conclusion, we assess the generalization capabilities of the fine-tuned model in \cref{subsec:y2024}.

\paragraph{Experimental setup}
\label{subsec:setup}

At inference time, SAM requires a prompt to produce the predictions. For a fair evaluation, we use the ``automatic'' of SAM, which consists of providing a regular grid of points ($32\times32$) as a prompt, then post-processing the predictions by eliminating masks with low confidence, and finally using non-maximal suppression to eliminate duplicates. As metrics, we report the mAP50, the main metric in instance segmentation, defined as the mean Average Precision computed at a fixed Intersection over Union (IoU) threshold of 0.5 (even though we have only one class, we still use the wording `mean' for clarity). In addition, we report the mAR150, defined as the mean Average Recall (AR) computed with a maximum of 150 detections per image.

\subsection{Zero-shot Assesment}
\label{subsec:zeroshot}
We first extensively evaluate the performance obtained in a zero-shot setup. The SAM authors provide 3 pre-trained models based on the backbones ViT-B (``Base'', $93M$ params), and ViT-L (``Large'', $312M$ params) ViT-H (``Huge'', $613M$ Params) \cite{dosovitskiy2020image}. The results are reported in \Cref{tab:sam-zs}. In this configuration, the results obtained are mediocre on all three datasets. Even with the most powerful backbones, we do not come close to acceptable performance. This is a symptom of the excessive domain shift between satellite imagery and the SAM training set, highlighting the need for fine-tuning on the target domain.

\subsection{Training on AI4B}
\label{subsec:sam_ft}
We fine-tune SAM using the LoRA method, focusing only on the ViT-B backbone.
The optimization setup follows the original SAM framework. SAM generates three possible masks for each prompt to better model the inherent ambiguity in mask definition. As in the original work, we compute the loss between the ground-truth mask and all three predictions, backpropagating the loss corresponding to the lowest value. For the IoU score prediction task \cite{kirillov2023segment}, the target is defined as the IoU between the current predicted mask and the true mask. We use the AdamW optimizer with weight decay $5e^{-4}$, with a linear warm-up of the learning rate to $5e^{-5}$ over $256$ steps, then held constant, a batch size of 4 across two GPUs.  %
Unlike other efficient fine-tuning schemes, we train the model at full 32-bit floating-point precision, having observed instability by combining this setup with reduced-precision computation.

\begin{table}[h]
\begin{minipage}[t]{0.47\textwidth}
\centering
\caption{Results with LoRA fine-tuning on SAM with the standard AI4B dataset. $r$ denotes the LoRA rank. TP denotes the number of Trainable Parameters.}
    \centering
    \scalebox{0.85}{
    \begin{tabular}{cccccccc} %
    \toprule
    \textbf{Decoder} & \textbf{Prompt} & \textbf{$r$} & $\textbf{TP}$ & \textbf{mAP50} & \textbf{mAR150} \\
    \midrule
    Frozen & Sing & $8$ & $294\text{K}$ & $25.5$ & $42.2$ \\
    Full & Sing & $8$ & $ 4.2\text{M}$ & $21.3$ & $41.0$ \\
    LoRA & Sing & $8$ & $ 342\text{K}$ & $25.5$ & $42.3$ \\ 
    LoRA & Mul & $8$ & $ 342\text{K}$ & $25.5$ & $42.9$ \\ 
    \midrule
    Frozen & Sing & $32$ & $1.2\text{M}$ & $25.5$ & $42.9$ \\ 
    LoRA & Sing & $32$ & $1.4\text{M}$ & $\mathbf{27.0}$ & $\mathbf{43.3}$ \\ 
    LoRA & Mul & $32$ & $1.4\text{M}$ & $26.2$ & $43.3$ \\ 
    \bottomrule
    
\end{tabular}
 }
\label{tab:ai4b-res}
  \end{minipage}
    \hfill
    \begin{minipage}[t]{0.47\textwidth}
    \caption{Zero-shot results with the original SAM weights, provided by the paper.}
    \centering
    \scalebox{0.8}{
    \begin{tabular}{lccc}
    \toprule
    \textbf{Dataset}  & \textbf{Backbone} & \textbf{mAP50($\uparrow$)} & \textbf{mAR150($\uparrow$)} \\
    \midrule
    
    AI4B & ViT-B & $6.6$ & $13.7$ \\
    AI4B & ViT-L & $5.9$ & $14.7$ \\ 
    AI4B & ViT-H & $5.7$  & $15.1$  \\ \midrule
    
    ERAS (S2) & ViT-B & $2.4$ & $4.8$ \\ 
    ERAS (S2) & ViT-L & $2.6$ & $5.5$ \\
    ERAS (S2) & ViT-H & $2.8$ & $6.1$ \\ \midrule
    
    ERAS (HR) & ViT-B & $9.1$ & $1.66$ \\
    ERAS (HR) & ViT-L & $9.6$ & $1.91$ \\
    ERAS (HR) & ViT-H & $10.5$ & $2.16$ \\
    
    \bottomrule
    \end{tabular}
     }
    \label{tab:sam-zs}
    \end{minipage}
\end{table}

\paragraph{Alternatives investigated}
As introduced in \Cref{subsec:sft}, we are interested in clarifying three key options: %
(i) how to finetune the mask decoder, respectively, frozen, fine-tuned entirely, or finetuned with LoRA, (b) how to define the prompt as a single point or multiple points, respectively, (c) evaluate the impact of LoRA rank. %
At first, we fix the LoRA rank \red{to the lowest reasonable level} ($r=8$) and the prompting strategy to a single point per mask and evaluate the three options for the mask decoder. \Cref{tab:ai4b-res} (lines 1-3) shows that the full training of the mask decoder yields the worst performance with the highest number of trainable parameters; conversely, applying LoRA to the decoder layers adds only 48K trainable parameters, with no apparent effect on performance. Switching from a single-point prompt to a multi-point prompt (lines 3-4) does not seem to affect the learning process either. We assume the very small number of trainable parameters is the bottleneck of learning, so (lines 5-7) we increase the rank $r$ to 32 and repeat the inconclusive experiments. Again, we observe that multi-point prompting does not improve performance while adding cost to data preparation. In contrast, in this case, we note that applying LoRA to the decoder has a positive effect on performance, at the cost of a small increase in parameter count. %

\subsection{Training on ERAgriSeg dataset}

Below, we analyze the results obtained by performing fine-tuning on the proposed ERAgriSeg (ERAS) datasets. The training follows the best setup identified in the Ai4B setting ($r$=32, LoRA applied to the decoder and prompting with a single point per instance mask), keeping the same choices of hyperparameter and optimizer. In addition, we have the option of training the LoRA weights from scratch or further fine-tuning the weights trained for AI4Boundary.

\begin{table}[h]
\begin{minipage}[t]{0.49\textwidth}
        \centering
    \caption{Results with specialized LoRA fine-tuning on different weights with different datasets.}
     \scalebox{0.8}{
    \centering
    \begin{tabular}{lcccc} %
    \toprule
        \textbf{Train} & \textbf{Pretrain} & \textbf{Test} & \textbf{mAP50} &\textbf{mAR150} \\
        \midrule
        ERAS(S2) & SAM & ERAS(S2) & 17.1 & 30.35 \\ 
        ERAS(S2) & AI4B & ERAS(S2) & 17.1 & 30.5\\ \midrule  
        ERAS(HR) & SAM & ERAS(HR) & 19.3 & 34.0 \\
        ERAS(HR) & AI4B & ERAS(HR) & 19.4 & 34.3 \\ 
        \bottomrule
    \end{tabular}
    }
    \label{tab:sam-res}
    \end{minipage}
    \hfill
    \begin{minipage}[t]{0.45\textwidth}
    \centering
        \caption{Results on Mask R-CNN and LoRA fine-tuning of SAM for segmentation masks on AI4B dataset. }
        \scalebox{0.85}{
    \begin{tabular}{ccc} 
    \toprule
        \textbf{Model}  & \textbf{Dataset} & \textbf{mAP50} \\
        \midrule
        Mask R-CNN  & AI4B & $14.6$ \\
        SAM (LoRA) & AI4B & $\mathbf{27.0}$ \\
        \midrule
        Mask R-CNN  & ERAS (S2) & $10.9$ \\
        SAM (LoRA) & ERAS (S2) & $\mathbf{17.1}$ \\
        \midrule
         Mask R-CNN  & ERAS (HR) & $11.8$ \\
        SAM (LoRA) & ERAS (HR) & $\mathbf{19.3}$ \\
        \bottomrule
    \end{tabular}
    \label{tab:rcnn-res}
    }
     \end{minipage}
\end{table}

\noindent The most surprising result is that two-step fine-tuning, by training the LoRA weight on AI4B first, does not seem to bring any improvement. We can track down this phenomenon to the small number of trainable parameters, implying a poor ability to retain previous knowledge. The mAP obtained on the HR dataset is slightly higher. Although a direct comparison is non-trivial, due to the different temporal distribution of the test images, \red{we can still assume a partial benefit brought by the higher resolution. }
\paragraph{Baseline Results}\label{subsec:maleresults}
\label{subsec:rcnn_baseline}

For comparison, we train the Mask-RCNN instance-segmentation model\cite{he2017mask}.
The configuration of choice is based on a ResNet-50 backbone with a Feature Pyramid Network (FPN) architecture. The model is initialized with pretrained ResNet-50 weights. The learning rate is set to $5e-5$ and weight decay to $0.01$, training with \red{Adam} optimizer and batch size of $8$.
The results of the comparison are shown in the table \Cref{tab:rcnn-res}, with SAM consistently outperforming Mask-RCNN on all the datasets. %

\subsection{Generalization to year 2024} \label{subsec:y2024}

The ERAS dataset used for all the previous experiments exclusively includes data for the year 2023 (both in train and in test). We have data for the first quarter of the year 2024, only for the Sentinel-2 mode, which we can use to further assess the generalization capabilities. For this experiment, instead, we use as the validation set all the images for Q1 of 2024, for all provinces.

\begin{table}[h]
    \caption{Assessment of generalization to the year 2024}
    \centering
     \scalebox{0.8}{
    \begin{tabular}{lcccccc} %
    \toprule
    \textbf{Train} & \textbf{Pretrain} & \textbf{Test} & \textbf{mAP50} & \textbf{mAR150} \\
    \bottomrule
    0-Shot   & -  & [2024-Q1] & $2.3$ & $4.3$ \\
    AI4B   & -  & [2024-Q1] & $12.9$ & $24.0$ \\  
    \midrule
    ERAS (S2)   & AI4B  & [2023] & $17.1$ & $30.5$ \\ 
    \midrule
    ERAS (S2)   & SAM  & [2024-Q1] & $16.1$ & $29.3$ \\ 
    ERAS (S2)  & AI4B  & [2024-Q1] & $\mathbf{16.2}$ & $29.4$ \\
    \bottomrule
\end{tabular}
}
\label{tab:y2024}
\end{table}
\noindent

\Cref{tab:y2024} reports results for SAM (Zero-shot), AI4B fine-tuned, and ERAS 2023 fine-tuned models, with and without AI4B pretraining. Results align closely with \Cref{tab:sam-res} (2023), confirming strong temporal generalization. Our AI4B-pretrained, ERAS(S2)-fine-tuned model shows only a $0.9$ drop.

\section{Conclusions}
\label{sec:concl}

In this paper, we introduced a robust fine-tuning pipeline to extend the state-of-the-art Segment Anything Model (SAM) to the field delineation task. The proposed pipeline vastly outperforms the baseline instance segmentation method, showcasing promising results for broader applications of vision foundation models in the Earth observation domain. The proposed regional dataset, ERAgriSeg (ERAS), has proven to be a valuable asset to assess the generalization ability to regions not covered by literature datasets. In addition, providing the ability to observe the implications of higher resolution images. We believe that these contributions will aid researchers and practitioners in developing more efficient and accurate models for satellite image analysis, ultimately leading to better-informed decisions in agricultural and environmental domains.

\section*{Acknowledgements}
\label{sec:ack}
This work has been partially supported by the industrial collaboration between ABACO Group SpA and UNIMORE. C. Scribano work was partly funded by the Partenariato Esteso PE00000013 - ``FAIR'', funded by the European Commission under the NextGeneration EU program, PNRR - M4C2 - Investimento 1.3.

\bibliographystyle{splncs04}
\bibliography{main}
\end{document}